\documentclass[]{spie}  

\usepackage{amsmath,amsfonts,amssymb}
\usepackage{graphicx}
\usepackage[colorlinks=true, allcolors=blue]{hyperref}

\usepackage{booktabs}
\usepackage{multirow}

\newcommand{\vt}[1]{{\boldsymbol #1}}

\title{Domain-invariant feature learning in brain MR imaging for content-based image retrieval}

\author[a]{Shuya Tobari}
\author[a]{Shuhei Tomoshige}
\author[a]{Hayato Muraki}
\author[b]{Kenichi Oishi}
\author[a]{Hitoshi Iyatomi}
\affil[ ]{for the Alzheimer's Disease Neuroimaging Initiative$^*$}
\affil[a]{Dept. of Science and Engineering, Hosei University,  3-7-2 Kajino Koganei, Tokyo, 184-8584, Japan}
\affil[b]{Dept. of Radiology and Radiological Science, Johns Hopkins Medicine, 225 Traylor Building, 720 Rutland Ave.Baltimore, MD 21205, USA}

\authorinfo{$^*$Data used in preparation of this article were obtained from the Alzheimer’s Disease Neuroimaging Initiative (ADNI) database (adni.loni.usc.edu). As such, the investigators within the ADNI contributed to the design and implementation of ADNI and/or provided data but did not participate in analysis or writing of this report. A complete listing of ADNI investigators can be found at: \url{http://adni.loni.usc.edu/wp-content/uploads/how_to_apply/ADNI_Acknowledgement_List.pdf}.\\
Send correspondence to H.Iyatomi: E-mail: iyatomi@hosei.ac.jp, Telephone: 81 42 387 6217}

\begin{document} 
\maketitle

\begin{abstract}
When conducting large-scale studies that collect brain MR images from multiple facilities, the impact of differences in imaging equipment and protocols at each site cannot be ignored, and this domain gap has become a significant issue in recent years.
In this study, we propose a new low-dimensional representation (LDR) acquisition method called style encoder adversarial domain adaptation (SE-ADA) to realize content-based image retrieval (CBIR) of brain MR images.
SE-ADA reduces domain differences while preserving pathological features by separating domain-specific information from LDR and minimizing domain differences using adversarial learning.
 In evaluation experiments comparing SE-ADA with recent domain harmonization methods on eight public brain MR datasets (ADNI1/2/3, OASIS1/2/3/4, PPMI), SE-ADA effectively removed domain information while preserving key aspects of the original brain structure and demonstrated the highest disease search accuracy.
\end{abstract}


\section{INTRODUCTION}
\label{sec:intro}  
The development of content-based image retrieval (CBIR) for the diagnosis and research reuse of accumulated magnetic resonance (MR) images is urgently needed \cite{kumar2013content}.
However, previous studies have identified a significant challenge posed by domain differences—biases introduced by variations in scanners, protocols, and other factors—which can significantly impact the consistency of images obtained from different imaging sites \cite{arai2021disease, wachinger2021detect, clark2006impact, onga2019efficient}.
To address these domain differences, methods such as ComBat \cite{johnson2007adjusting}, which uses empirical Bayesian techniques to adjust for variability between different data batches, have been widely adopted \cite{wachinger2021detect, fortin2018harmonization, pomponio2020harmonization, yu2018statistical}. 
While effective for aligning data distributions, these statistical methods are limited to situations where multiple target datasets are available, and they cannot harmonize brain images from unknown domains, making them unsuitable for CBIR.
In the context of deep learning-based harmonization of brain MR images, Dinsdale et al. \cite{dinsdale2021deep} demonstrated that incorporating the Adversarial Domain Adaptation (ADA) technique into a 3D convolutional autoencoder (3D-CAE) can harmonize MR images across different domains.
However, the challenge remains that achieving a low-dimensional representation (LDR) that retains comprehensive brain information while simultaneously removing domain differences involves conflicting learning objectives, leaving room for improvement in accuracy with the ADA approach.
To address these limitations, Tobari et al. \cite{tobari2023} proposed the multi-decoder adversarial domain adaptation (MD-ADA) method. 
This approach extends ADA by equipping each domain with its own decoder, thereby enhancing the domain harmonization capability and achieving more stable learning. 
However, the need for a separate decoder for each domain limits its scalability.
In this paper, we introduce a novel method called style encoder adversarial domain adaptation (SE-ADA) for more effectively acquiring domain-invariant LDRs. 
Unlike MD-ADA, which requires a separate decoder for each domain, SE-ADA utilizes a specialized style encoder that isolates domain information from the LDR, thereby eliminating domain-specific information while preserving the essential features for CBIR.

\section{SE-ADA: Style encoder adversarial domain adaptation}
\label{sec:SE-ADA}
\subsection{Prerequisite}
In the CBIR system we are developing, input 3D brain MRI images are compressed to a lower dimension using a feature extractor and compared with similarly compressed images in a database.
The goal of this research is to perform similarity calculations using LDRs that retain essential disease detection information while eliminating domain-specific biases, ensuring accurate and relevant search results.

\subsection{Structure and overview}

Figure 1 illustrates the structure of SE-ADA. 
SE-ADA builds on the adversarial domain adaptation (ADA) framework \cite{ganin2016domain, dinsdale2021deep}, integrating a domain predictor $g_D$ into a 3D convolutional autoencoder (3D-CAE).
SE-ADA introduces a new Style encoder $f_{SE}$, separate from the main encoder, which generates a LDR $\vt{z}$ of the original data $\vt{x}$.
This style encoder extracts only the style information $\vt{z^{(k)}}$ of the domain $k$ to which $\vt{x}$ belongs.
This style encoder is designed to extract only the domain-specific style information $\vt{z^{(k)}}$ necessary for image reconstruction.
The decoder $f_D$ reconstructs images using the sum of $\vt{z}$ and $\vt{z^{(k)}}$. 
In other words, by training the model so that the LDR $\vt{z}$ generated by the primary encoder is devoid of domain information, SE-ADA enables the reconstruction of images that retain domain characteristics, even when the LDR $\vt{z}$ is entirely domain-agnostic.

\begin{figure*}[ht]
\begin{center}
\includegraphics[width=170mm]{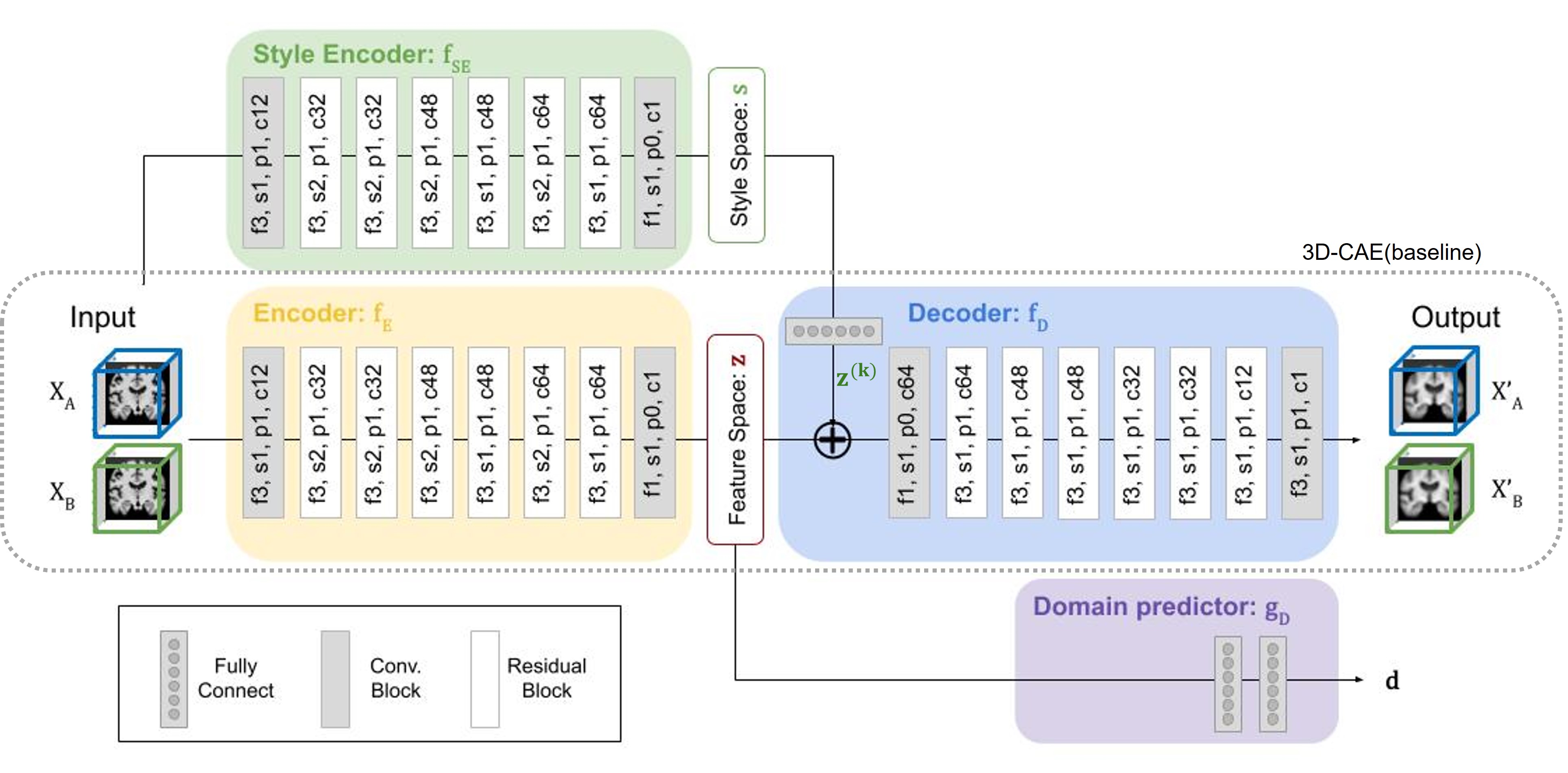}
\caption{Architecture of SE-ADA.}
\end{center}
\end{figure*}

\subsection{Training}
For the training data used in SE-ADA, the input image $\vt{x}$ is used to generate the reconstructed image $\vt{x'}$, the output $\vt{d}$ from the domain predictor $g_D$, and the one-hot domain label corresponding to the output $\vt{s}$ from the style encoder $f_{SE}$.
To enhance the effectiveness of $z^{(k)}$ as a representation, the output $\vt{s}$ from the style encoder is initially trained through supervised learning using the domain label.
Subsequently, $\vt{z^{(k)}}$ is refined through a fully connected network based on this training.
The training of SE-ADA is conducted by alternately repeating the following three stages. In each stage, components not explicitly mentioned remain with fixed weights and are not updated.

\begin{description}
\item[Stage 1]
The Encoder $f_E$, Decoder $f_D$, and Style Encoder $f_{SE}$ are trained to minimize the reconstruction error and style error for $\vt{s}$, resulting in the acquisition of updated LDR $\vt{z}$, styles $\vt{s}$, and $\vt{z^{(k)}}$.
\item[Stage 2]
The Domain predictor $g_D$ is trained to accurately classify domains based on the LDR $\vt{z}$.
\item[Stage 3]
The Encoder $f_E$ is trained to prevent domain classification from the LDR $\vt{z}$ while keeping $g_D$ fixed.
\end{description}

To optimize the different loss functions associated with each stage, training is conducted by repeating stages 1 through 3 within each training batch.
The removal of domain information is achieved through the combined training in stages 2 and 3.
However, during stage 3, when performing harmonic learning of domain differences, training is conducted using only data from healthy individuals (CN) to ensure that differences due to pathological features are not eliminated from the acquired LDRs.

\section{Experiments}
\subsection{Data and pre-processing}
Table 1 summarizes the datasets used in the experiment. As shown, this experiment utilized eight datasets from the Alzheimer's Disease Neuroimaging Initiative (ADNI) 1/2/3 \cite{mueller2005alzheimer}, the Open Access Series of Imaging Studies (OASIS) 1/2/3/4 \cite{marcus2007open, koenig2020select}, and the Parkinson's Progression Markers Initiative (PPMI) (\url{www.ppmi-info.org/data}).
The ADNI was launched in 2003 as a public-private partnership, led by Principal Investigator Michael W. Weiner, MD. The primary goal of ADNI has been to test whether serial MRI, positron emission tomography (PET), other biological markers, and clinical and neuropsychological assessments can be combined to measure the progression of mild cognitive impairment (MCI) and early Alzheimer’s disease (AD). For up-to-date information, see \url{www.adni-info.org}.
The experiment included three types of case image data: cognitively normal (CN), Alzheimer's disease (AD), and mild cognitive impairment (MCI).
A total of 10 different domains were tested, treating each magnetic field strength as a distinct domain.
The columns labeled "train" and "test" in the table indicate the domain data used for training and evaluating the model, respectively.
For all MRI images used in the experiments, skull stripping was performed using OpenMAP-T1 \cite{nishimaki2024openmap} followed by affine transformation (including translation and rotation) for alignment.
The resolution was then standardized to 2 mm cubic voxels, and the image size was reduced to 80$\times$112$\times$80 pixels. Additionally, for brightness value normalization, negative values and values exceeding $4\sigma$ were replaced with 0 and the $4\sigma$ value, respectively, followed by linear conversion to a range of [0,1].

\begin{table}[ht]
\caption{Data used in this study: The numbers in parentheses indicate the number of patients.}
\centering
\label{tab:dataset}
\begin{tabular}{@{}crrrrrcc@{}}
\toprule
dataset                 & \multicolumn{1}{c}{strength [T]} & \multicolumn{1}{c}{CN} & \multicolumn{1}{c}{AD} & \multicolumn{1}{c}{MCI} & \multicolumn{1}{c}{total} & train                & test                 \\ \midrule
\multirow{2}{*}{ADNI1} & 1.5                      & 480 (171)               & 490 (227)               & 691(267)                & 1661 (570)              & $\checkmark$                    & \multicolumn{1}{l}{} \\
                       & 3.0                      & 100 (36)                & 96 (52)                 & -                       & 196 (88)                & \multicolumn{1}{l}{} & $\checkmark$                    \\
\multirow{2}{*}{ADNI2} & 1.5                      & 338 (78)                & 243 (68)                & 281 (75)                 & 862 (188)               & $\checkmark$                    & \multicolumn{1}{l}{} \\
                       & 3.0                      & 2367 (337)              & 1190 (238)              & 3425 (502)               & 6982 (930)              & $\checkmark$                    & \multicolumn{1}{l}{} \\
ADNI3                  & 3.0                      & 990 (460)               & 213 (105)               & 658 (284)                & 1861 (786)              & $\checkmark$                    & \multicolumn{1}{l}{} \\
OASIS1                 & 1.5                      & 526 (135)               & 387 (100)               & -                       & 913 (235)               & \multicolumn{1}{l}{} & $\checkmark$                    \\
OASIS2                 & 1.5                      & 826 (86)                & 542 (64)                & -                       & 1368 (150)              & $\checkmark$                    & \multicolumn{1}{l}{} \\
OASIS3                 & 3.0                      & 518 (412)               & 90 (84)                 & -                       & 608 (492)               & \multicolumn{1}{l}{} & $\checkmark$                    \\
OASIS4                 & 3.0                      & 42 (42)                 & 191 (191)               & -                       & 233 (233)                   & \multicolumn{1}{l}{} & $\checkmark$                    \\
PPMI                   & 3.0                      & 183 (112)               & -                      & -                       & 183 (112)               & \multicolumn{1}{l}{} & $\checkmark$                    \\ \bottomrule
\end{tabular}
\end{table}

\subsection{Evaluations}
In order to verify the effectiveness of the LDR $\vt{z}$ obtained by SE-ADA for CBIR, we evaluated it using the following four criteria: A) Brain image preservation (RMSE, SSIM), B) Alzheimer's disease diagnostic performance (Diag. F1), C) Domain prediction performance (Domain F1), and D) Versatility for unknown domain data (Clustering).
%
%
For criterion (C), a score close to 1/5, which represents the reciprocal of the number of domains, indicates high performance in harmonizing domains.
Criterion (D) was assessed by evaluating the extent to which the variance in CN data was reduced through harmonization.
This was measured using six representative clustering indices: silhouette score, homogeneity score, completeness score, V-measure, adjusted Rand index, and adjusted mutual information.
The results were expressed as the average reduction from the baseline CAE.

The LDRs obtained from the 3D-CAE model as shown in Figure 1 were harmonized using ComBat \cite{johnson2007adjusting}, ADA \cite{dinsdale2021deep}, and MD-ADA \cite{tobari2023}, which were implemented for comparison.
MD-ADA, a successor to ADA, enhances performance by incorporating multiple Decoders to match the number of domains.
As described in the original paper, the multiple decoder implementation is designed to minimize the number of parameters by branching the fully connected layer in the final part, reducing the likelihood of overfitting.
Both ADA and MD-ADA follow the same three-stage training process as SE-ADA, with the exception that they do not include a Style Encoder.
As mentioned earlier, ComBat aligns the statistics of each data group, making it unsuitable for CBIR, as it cannot handle practical tasks like dimensional compression of brain images from unknown domains and searching for similar cases. For comparison, the results of adding Gaussian noise $N(0,0.1)$ to the acquired LDR are also presented.

\section{Results and Discussion}
Table 2 presents the results of the comparison for each evaluation method. 
Noise and ComBat are methods that modify the acquired feature representations directly, and since they do not generate reconstructed images that can be fairly evaluated, these methods were not included in the evaluation.
Additionally, the results for ComBat are provided for reference only, as it cannot be harmonized with a single set of execution data.

A significant deviation from the original result would indicate that the model is highly influenced by domain differences.
It can be observed that the CAE (baseline) has limited diagnostic capability for diseases in unknown domain data.
ADA was able to reduce domain classification accuracy to some extent through adversarial learning, but this came at the cost of significantly reduced diagnostic capability for diseases and diminished preservation of brain structure.
This incompatibility arises because adversarial domain adaptation and CAE learning are not well-aligned.
MD-ADA addresses some of ADA's shortcomings by incorporating a decoder for each domain, resulting in notable improvements in both the preservation of brain structure and disease diagnostic capability. However, its performance in domain harmonization remains insufficient.
The proposed SE-ADA method achieves comparable results to other methods in terms of preserving brain information while demonstrating the highest diagnostic performance for diseases in unknown domain data.
Additionally, SE-ADA shows superior domain harmonization, acquiring a desirable LDR that resists classification by domain.
This is evidenced by the minimal difference in diagnostic performance between data within the same domain and data from unknown domains after harmonization, as well as by its excellent clustering performance.

\begin{table}[ht]
\caption{Summary of evaluation results for the obtained LDR.}
\label{tab:results_table}
\centering
\begin{tabular}{@{}lrrrrr@{}}
\toprule
         & \multicolumn{2}{c}{Data preservation} & \multicolumn{1}{c}{Diag. capability} & \multicolumn{2}{c}{Domain harmonization (5 domains)} \\ 
         & \multicolumn{1}{c}{RMSE $\downarrow$} & \multicolumn{1}{c}{SSIM $\uparrow$} & \multicolumn{1}{c}{Diag. F1 $\uparrow$ $\dag$} & \multicolumn{1}{c}{Domain F1 $\downarrow$ $\ddag$}  & \multicolumn{1}{c}{Clustering [$\%$]  $\downarrow$ } \\ \midrule
3D-CAE (baseline)     & \textbf{0.0764}$\pm$0.0307 & \textbf{0.873}$\pm$0.019 & 0.669 (0.807) & 0.631 & 0 \\
+ Noise  & \multicolumn{2}{c}{n/a}              & 0.648 (0.765) & 0.515 & -5.02 \\
+ Combat \cite{johnson2007adjusting} & \multicolumn{2}{c}{n/a}              & 0.713 (0.816) & 0.145 & -90.00 \\
+ ADA \cite{dinsdale2021deep}   & 0.0902$\pm$0.0359 & 0.817$\pm$0.023 & 0.664 (0.761) & 0.468 & -67.53 \\
+ MD-ADA \cite{tobari2023} & 0.0780$\pm$0.0319 & 0.868$\pm$0.021 & 0.717 (0.810) & 0.450 & -78.86 \\
\textbf{+ SE-ADA} & 0.0885$\pm$0.0357 & 0.843$\pm$0.023 & \textbf{0.747} (0.776) & \textbf{0.190} & \textbf{-85.19} \\ \bottomrule
\end{tabular}
\flushleft
${}^\dag$ The numbers in parentheses represent the diagnostic results within the same domain when the training data is split into an 8:2 ratio based on the patients. The large difference between the results inside and outside the parentheses indicates that the model is strongly affected by domain differences.\\
${}^\ddag$ A value close to 1/5 indicates that the domain cannot be estimated from $\vt{z}$, which is the desired outcome.\\
\end{table}

\section{Conclusion}
In this paper, we propose a new and effective method called style encoder adversarial domain adaptation (SE-ADA) for achieving CBIR of brain MR images.
SE-ADA introduces a style encoder that allows the retention of original brain structure features in the required LDR while eliminating only domain-related information. This approach achieves superior disease diagnostic performance for cases in unknown domains compared to other advanced methods.
As research based on large-scale data from multiple locations becomes increasingly important, SE-ADA offers an effective solution for acquiring LDR, particularly for CBIR applications.

\section{Acknowledgments}
This research was supported in part by the Ministry of Education, Science, Sports and Culture of Japan (JSPS KAKENHI), Grant-in-Aid for Scientific Research (C), 21K12656 (2021-2024) and 24K17506 (2024-2027).  \\
Data used in preparation of this article were obtained from the Alzheimer's Disease Neuroimaging Initiative (ADNI) database (adni.loni.usc.edu).
As such, the investigators within the ADNI contributed to the design and implementation of ADNI and/or provided data but did not participate in analysis or writing of this report. A complete listing of ADNI investigators can be found at: \url{http://adni.loni.usc.edu/wp-content/uploads/how\_to\_apply/ADNI\_Acknowledgement\_List.pdf}.\\
Data were provided by OASIS1/2/3/4 Cross-Sectional: Principal Investigators: D. Marcus, R, Buckner, J, Csernansky J. Morris; P50 AG05681, P01 AG03991, P01 AG026276, R01 AG021910, P20 MH071616, U24 RR021382. Longitudinal: Principal Investigators: D. Marcus, R, Buckner, J. Csernansky, J. Morris; P50 AG05681, P01 AG03991, P01 AG026276, R01 AG021910, P20 MH071616, U24 RR021382. Longitudinal Multimodal Neuroimaging: Principal Investigators: T. Benzinger, D. Marcus, J. Morris; NIH P30 AG066444, P50 AG00561, P30 NS09857781, P01 AG026276, P01 AG003991, R01 AG043434, UL1 TR000448, R01 EB009352. AV-45 doses were provided by Avid Radiopharmaceuticals, a wholly owned subsidiary of Eli Lilly, and Clinical Cohort: Principal Investigators: T. Benzinger, L. Koenig, P. LaMontagne.\\
An additional MRI data used in the preparation of this article were obtained from the Parkinson’s Progression Markers Initiative (PPMI) database (\url{www.ppmi-info.org/data}).
For up-to-date information on the study, visit www.ppmi-info.org. PPMI – a public-private partnership – is funded by the Michael J. Fox Foundation for Parkinson’s Research and funding partners, including AbbVie, Allergan, Avid Radiopharmaceuticals, Biogen, Biolegend, Bristol-Myers Squibb, Celgene, Denali, GE Healthcare, Genentech, GlaxoSmithKline, Lilly, Lundbeck, Merck, Meso Scale Discovery, Pfizer, Piramal, Prevail Therapeutics, Roche, Sanofi Genzyme, Servier, Takeda, Teva, UCB, Verily, Voyager Therapeutics, and Golub Capital.



\bibliography{ref_bali} 

\begin{thebibliography}{10}

\bibitem{kumar2013content}
Kumar, A., Kim, J., Cai, W., Fulham, M., and Feng, D., ``{Content-based medical image retrieval: a survey of applications to multidimensional and multimodality data},'' {\em Journal of digital imaging}~{\bf 26}(6),  1025--1039 (2013).

\bibitem{arai2021disease}
Arai, H., Onga, Y., Ikuta, K., Chayama, Y., Iyatomi, H., and Oishi, K., ``{Disease-oriented image embedding with pseudo-scanner standardization for content-based image retrieval on 3D brain MRI},'' {\em IEEE Access}~{\bf 9},  165326--165340 (2021).

\bibitem{wachinger2021detect}
Wachinger, C., Rieckmann, A., P{\"o}lsterl, S., Initiative, A. D.~N., et~al., ``{Detect and correct bias in multi-site neuroimaging datasets},'' {\em Medical Image Analysis}~{\bf 67},  101879 (2021).

\bibitem{clark2006impact}
Clark, K.~A., Woods, R.~P., Rottenberg, D.~A., Toga, A.~W., and Mazziotta, J.~C., ``{Impact of acquisition protocols and processing streams on tissue segmentation of T1 weighted MR images},'' {\em NeuroImage}~{\bf 29}(1),  185--202 (2006).

\bibitem{onga2019efficient}
Onga, Y., Fujiyama, S., Arai, H., Chayama, Y., Iyatomi, H., and Oishi, K., ``{Efficient feature embedding of 3D brain MRI images for content-based image retrieval with deep metric learning},'' in [{\em 2019 IEEE International Conference on Big Data (Big Data)}{\nolinebreak\hspace{0.1em}]},   3764--3769, IEEE (2019).

\bibitem{johnson2007adjusting}
Johnson, W.~E., Li, C., and Rabinovic, A., ``{Adjusting batch effects in microarray expression data using empirical Bayes methods},'' {\em Biostatistics}~{\bf 8}(1),  118--127 (2007).

\bibitem{fortin2018harmonization}
Fortin, J.-P., Cullen, N., Sheline, Y.~I., Taylor, W.~D., Aselcioglu, I., Cook, P.~A., Adams, P., Cooper, C., Fava, M., McGrath, P.~J., et~al., ``{Harmonization of cortical thickness measurements across scanners and sites},'' {\em Neuroimage}~{\bf 167},  104--120 (2018).

\bibitem{pomponio2020harmonization}
Pomponio, R., Erus, G., Habes, M., Doshi, J., Srinivasan, D., Mamourian, E., Bashyam, V., Nasrallah, I.~M., Satterthwaite, T.~D., Fan, Y., et~al., ``{Harmonization of large MRI datasets for the analysis of brain imaging patterns throughout the lifespan},'' {\em NeuroImage}~{\bf 208},  116450 (2020).

\bibitem{yu2018statistical}
Yu, M., Linn, K.~A., Cook, P.~A., Phillips, M.~L., McInnis, M., Fava, M., Trivedi, M.~H., Weissman, M.~M., Shinohara, R.~T., and Sheline, Y.~I., ``{Statistical harmonization corrects site effects in functional connectivity measurements from multi-site fMRI data},'' {\em Human brain mapping}~{\bf 39}(11),  4213--4227 (2018).

\bibitem{dinsdale2021deep}
Dinsdale, N.~K., Jenkinson, M., and Namburete, A.~I., ``{Deep learning-based unlearning of dataset bias for MRI harmonisation and confound removal},'' {\em NeuroImage}~{\bf 228},  117689 (2021).

\bibitem{tobari2023}
Tobari, S., Oishi, K., and Iyatomi, H., ``Acquiring a low-dimensional, environment-independent representation of brain {MR} images for content-based image retrieval,'' in [{\em {IEEE} International Conference on Systems, Man, and Cybernetics, {SMC} 2023, Honolulu, Oahu, HI, USA, October 1-4, 2023}{\nolinebreak\hspace{0.1em}]},   5096--5101, {IEEE} (2023).

\bibitem{ganin2016domain}
Ganin, Y., Ustinova, E., Ajakan, H., Germain, P., Larochelle, H., Laviolette, F., Marchand, M., and Lempitsky, V., ``{Domain-adversarial training of neural networks},'' {\em The journal of machine learning research}~{\bf 17}(1),  2096--2030 (2016).

\bibitem{mueller2005alzheimer}
Mueller, S.~G., Weiner, M.~W., Thal, L.~J., Petersen, R.~C., Jack, C., Jagust, W., Trojanowski, J.~Q., Toga, A.~W., and Beckett, L., ``{The Alzheimer's disease neuroimaging initiative},'' {\em Neuroimaging Clinics}~{\bf 15}(4),  869--877 (2005).

\bibitem{marcus2007open}
Marcus, D.~S., Wang, T.~H., Parker, J., Csernansky, J.~G., Morris, J.~C., and Buckner, R.~L., ``{Open Access Series of Imaging Studies (OASIS): cross-sectional MRI data in young, middle aged, nondemented, and demented older adults},'' {\em Journal of cognitive neuroscience}~{\bf 19}(9),  1498--1507 (2007).

\bibitem{koenig2020select}
Koenig, L.~N., Day, G.~S., Salter, A., Keefe, S., Marple, L.~M., Long, J., LaMontagne, P., Massoumzadeh, P., Snider, B.~J., Kanthamneni, M., et~al., ``{Select Atrophied Regions in Alzheimer disease (SARA): An improved volumetric model for identifying Alzheimer disease dementia},'' {\em NeuroImage: Clinical}~{\bf 26},  102248 (2020).

\bibitem{nishimaki2024openmap}
Nishimaki, K., Onda, K., Ikuta, K., Uchida, Y., Mori, S., Iyatomi, H., and Oishi, K., ``{OpenMAP-T1: A Rapid Deep Learning Approach to Parcellate 280 Anatomical Regions to Cover the Whole Brain},'' {\em medRxiv} ,  2024--01 (2024).

\end{thebibliography}
\bibliographystyle{spiebib} 

\end{document}